# ITEM2VEC: NEURAL ITEM EMBEDDING FOR COLLABORATIVE FILTERING

*Oren Barkan^\* and Noam Koenigstein\**

^Tel Aviv University
\*Microsoft

## ABSTRACT

Many Collaborative Filtering (CF) algorithms are item-based in the sense that they analyze item-item relations in order to produce item similarities. Recently, several works in the field of Natural Language Processing (NLP) suggested to learn a latent representation of words using neural embedding algorithms. Among them, the Skip-gram with Negative Sampling (SGNS), also known as word2vec, was shown to provide state-of-the-art results on various linguistics tasks. In this paper, we show that item-based CF can be cast in the same framework of neural word embedding. Inspired by SGNS, we describe a method we name item2vec for item-based CF that produces embedding for items in a latent space. The method is capable of inferring item-item relations even when user information is not available. We present experimental results that demonstrate the effectiveness of the item2vec method and show it is competitive with SVD.

## 1. INTRODUCTION AND RELATED WORK

Computing item similarities is a key building block in modern recommender systems. While many recommendation algorithms are focused on learning a low dimensional embedding of users and items simultaneously [1, 2, 3], computing item similarities is an end in itself. Item similarities are extensively used by online retailers for many different recommendation tasks. This paper deals with the overlooked task of learning item similarities by embedding items in a low dimensional space.

Item-based similarities are used by online retailers for recommendations based on a single item. For example, in the Windows 10 App Store, the details page of each app or game includes a list of other similar apps titled "People also like". This list can be extended to a full page recommendation list of items similar to the original app as shown in Fig. 1. Similar recommendation lists which are based merely on similarities to a single item exist in most online stores e.g., Amazon, Netflix, Google Play, iTunes store and many others.

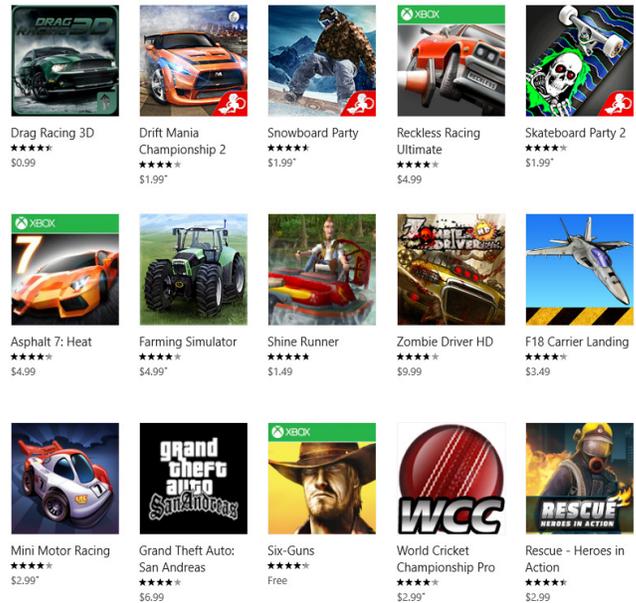

**Fig. 1**. Recommendations in Windows 10 Store based on similar items to Need For Speed.

The single item recommendations are different than the more "traditional" user-to-item recommendations because they are usually shown in the context of an explicit user interest in a specific item and in the context of an explicit user intent to purchase. Therefore, single item recommendations based on item similarities often have higher Click-Through Rates (CTR) than user-to-item recommendations and consequently responsible for a larger share of sales or revenue.

Single item recommendations based on item similarities are used also for a variety of other recommendation tasks: In "candy rank" recommendations for similar items (usually of lower price) are suggested at the check-out page right before the payment. In "bundle" recommendations a set of several items are grouped and recommended together. Finally, item similarities are used in online stores for better exploration and discovery and improve the overall user experience. It is unlikely that a user-item CF method, that learns the connections between items implicitly by defining slack variables for users, would produce better item representations than a method that is optimized to learn the item relations directly.

Item similarities are also at the heart of item-based CF algorithms that aim at learning the representation directly from the item-item relations [4, 5]. There are several scenarios where item-based CF methods are desired: in a large scale dataset, when the number of users is significantly larger than the number of items, the computational complexity of methods that model items solely is significantly lower than methods that model both users and items simultaneously. For example, online music services may have hundreds of millions of enrolled users with just tens of thousands of artists (items).

In certain scenarios, the user-item relations are not available. For instance, a significant portion of today's online shopping is done without an explicit user identification process. Instead, the available information is per session. Treating these sessions as "users" would be prohibitively expensive as well as less informative.

Recent progress in neural embedding methods for linguistic tasks have dramatically advanced state-of-the-art NLP capabilities [6, 7, 8, 12]. These methods attempt to map words and phrases to a low dimensional vector space that captures semantic relations between words. Specifically, Skip-gram with Negative Sampling (SGNS), known also as word2vec [8], set new records in various NLP tasks [7, 8] and its applications have been extended to other domains beyond NLP [9, 10].

In this paper, we propose to apply SGNS to item-based CF. Motivated by its great success in other domains, we suggest that SGNS with minor modifications may capture the relations between different items in collaborative filtering datasets. To this end, we propose a modified version of SGNS named item2vec. We show that item2vec can induce a similarity measure that is competitive with an item-based CF using SVD, while leaving the comparison to other more complex methods to a future research.

The rest of the paper is organized as follows: Section 2 overviews the SGNS method. Section 3 describes how to apply SGNS to item-based CF. In Section 4, we describe the experimental setup and present qualitative and quantitative results.

## 2. SKIP-GRAM WITH NEGATIVE SAMPLING

Skip-gram with negative sampling (SGNS) is a neural word embedding method that was introduced by Mikolov et. al in [8]. The method aims at finding words representation that captures the relation between a word to its surrounding words in a sentence. In the rest of this section, we provide a brief overview of the SGNS method.

Given a sequence of words $(w_i)_{i=1}^{K}$ from a finite vocabulary $W = \{w_i\}_{i=1}^{W}$, the Skip-gram objective aims at maximizing the following term:

$$\frac{1}{K}\sum_{i=1}^{K}\sum_{-c\leq j\leq c, j\neq 0}\log p(w_{i+j}|w_i) \quad (1)$$

where $c$ is the context window size (that may depend on $w_i$) and $p(w_j|w_i)$ is the softmax function:

$$p(w_j|w_i) = \frac{\exp(u_i^T v_j)}{\sum_{k\in I_W}\exp(u_i^T v_k)} \quad (2)$$

where $u_i \in U(\subset \mathbb{R}^m)$ and $v_i \in V(\subset \mathbb{R}^m)$ are latent vectors that correspond to the target and context representations for the word $w_i \in W$, respectively, $I_W \triangleq \{1,...,|W|\}$ and the parameter $m$ is chosen empirically and according to the size of the dataset.

Using Eq. (2) is impractical due to the computational complexity of $\nabla p(w_j|w_i)$, which is a linear function of the vocabulary size $|W|$ that is usually in size of $10^5 - 10^6$.

Negative sampling comes to alleviate the above computational problem by the replacement of the softmax function from Eq.(2) with

$$p(w_j|w_i) = \sigma(u_i^T v_j)\prod_{k=1}^{N}\sigma(-u_i^T v_k)$$

where $\sigma(x) = 1/1+\exp(-x)$, $N$ is a parameter that determines the number of negative examples to be drawn per a positive example. A negative word $w_i$ is sampled from the unigram distribution raised to the 3/4rd power. This distribution was found to significantly outperform the unigram distribution, empirically [8].

In order to overcome the imbalance between rare and frequent words the following subsampling procedure is proposed [8]: Given the input word sequence, we discard each word $w$ with a probability $p(discard | w) = 1 - \sqrt{\frac{\rho}{f(w)}}$ where $f(w)$ is the frequency of the word $w$ and $\rho$ is a prescribed threshold. This procedure was reported to accelerate the learning process and to improve the representation of rare words significantly [8].

Finally, $U$ and $V$ are estimated by applying stochastic gradient ascent with respect to the objective in Eq. (1).

## 3. ITEM2VEC – SGNS FOR ITEM SIMILARITY

In the context of CF data, the items are given as user generated sets. Note that the information about the relation between a user and a set of items is not always available. For example, we might be given a dataset of orders that a store received, without the information about the user that made the order. In other words, there are scenarios where multiple sets of items might belong to the same user, but this information is not provided. In Section 4, we present experimental results that show that our method handles these scenarios as well.

We propose to apply SGNS to item-based CF. The application of SGNS to CF data is straightforward once we realize that a sequence of words is equivalent to a set or basket of items. Therefore, from now on, we will use the terms "word" and "item" interchangeably.

By moving from sequences to sets, the spatial / time information is lost. We choose to discard this information, since in this paper, we assume a static environment where items that share the same set are considered similar, no matter in what order / time they were generated by the user. This assumption may not hold in other scenarios, but we keep the treatment of these scenarios out of scope of this paper.

Since we ignore the spatial information, we treat each pair of items that share the same set as a positive example. This implies a window size that is determined from the set size. Specifically, for a given set of items, the objective from Eq. (1) is modified as follows:

$$\frac{1}{K}\sum_{i=1}^{K}\sum_{j \neq i}^{K} \log p(w_j | w_i).$$

Another option is to keep the objective in Eq. (1) as is, and shuffle each set of items during runtime. In our experiments we observed that both options perform the same.

The rest of the process remains identical to the method described in Section 2. We name the described method item2vec.

In this work, we used $u_i$ as the final representation for the $i$-th item and the affinity between a pair of items is computed by the cosine similarity. Other options are to use $v_i$, the additive composition, $u_i + v_i$ or the concatenation $\left[ u_i^T \ v_i^T \right]^T$. Note that the last two options sometimes produce superior representation.

## 4. EXPERIMENTAL SETUP AND RESULTS

In this section, we present an empirical evaluation of the item2vec method. We provide both qualitative and quantitative results depending whether a metadata about the items exists. As a baseline item-based CF algorithm we used item-item SVD.

### 4.1 Datasets

We evaluate the methods on two different datasets, both private. The first dataset is user-artist data that is retrieved from the Microsoft Xbox Music service. This dataset consist of 9M events. Each event consists of a user-artist relation, which means the user played a song by the specific artist. The dataset contains 732K users and 49K distinct artists.

The second dataset contains orders of products from Microsoft Store. An order is given by a basket of items without any information about the user that made it. Therefore, the information in this dataset is weaker in the sense that we cannot bind between users and items. The dataset consist of 379K orders (that contains more than a single item) and 1706 distinct items.

### 4.2 Systems and parameters

We applied item2vec to both datasets. The optimization is done by stochastic gradient decent. We ran the algorithm for 20 epochs. We set the negative sampling value to $N = 15$ for both datasets. The dimension parameter $m$ was set to 100 and 40 for the Music and Store datasets, respectively. We further applied subsampling with $\rho$ values of $10^{-5}$ and $10^{-3}$ to the Music and Store datasets, respectively. The reason we set different parameter values is due to different sizes of the datasets.

We compare our method to a SVD based item-item similarity system. To this end, we apply SVD to a square matrix in size of number of items, where the $(i, j)$ entry contains the number of times $(w_i, w_j)$ appears as a positive pair in the dataset. Then, we normalized each entry according to the square root of the product of its row and column sums. Finally, the latent representation is given by the rows of $US^{1/2}$, where $S$ is a diagonal matrix that its diagonal contains the top $m$ singular values and $U$ is a matrix that contains the corresponding left singular vectors as columns. The affinity between items is computed by cosine similarity of their representations. Throughout this section we name this method "SVD".

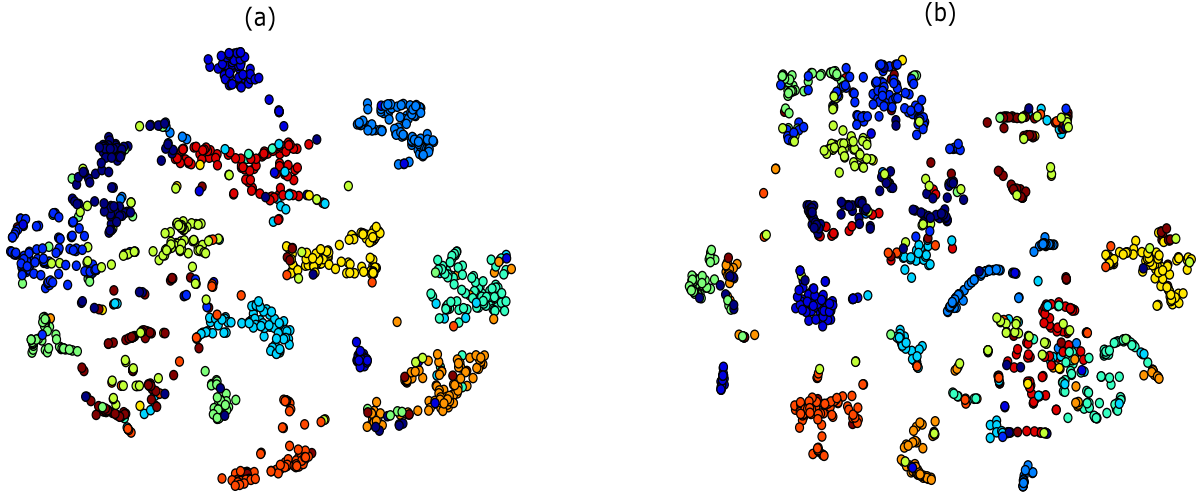

**Fig.2**: t-SNE embedding for the item vectors produced by item2vec (a) and SVD (b).
The items are colored according to a web retrieved genre metadata.

### 4.3 Experiments and results

The music dataset does not provide genre metadata. Therefore, for each artist we retrieved the genre metadata from the web to form a genre-artist catalog. Then we used this catalog in order to visualize the relation between the learnt representation and the genres. This is motivated by the assumption that a useful representation would cluster artists according to their genre. To this end, we generated a subset that contains the top 100 popular artists per genre for the following distinct genres: 'R&B / Soul', 'Kids', 'Classical', 'Country', 'Electronic / Dance', 'Jazz', 'Latin', 'Hip Hop', 'Reggae / Dancehall', 'Rock', 'World', 'Christian / Gospel' and 'Blues / Folk'. We applied t-SNE [11] with a cosine kernel to reduce the dimensionality of the item vectors to 2. Then, we colored each artist point according to its genre.

Figures 2(a) and 2(b) present the 2D embedding that was produced by t-SNE, for item2vec and SVD, respectively. As we can see, item2vec provides a better clustering. We further observe that some of the relatively homogenous areas in Fig. 2(a) are contaminated with items that are colored differently. We found out that many of these cases originate in artists that were mislabeled in the web or have a mixed genre.

Table 1 presents several examples, where the genre associated with a given artist (according to metadata that we retrieved from the web) is inaccurate or at least inconsistent with Wikipedia. Therefore, we conclude that usage based models such as item2vec may be useful for the detection of mislabeled data and even

**TABLE 1**: INCONSISTENCIES BETWEEN GENRES FROM THE WEB CATALOG AND THE ITEM2VEC BASED KNN PREDICTIONS

| Artist name | Genre from web catalog (incorrect) | Genre predicted by item2vec based Knn (correct) |
|---|---|---|
| DMX | R&B / Soul | Hip Hop |
| LLJ | Rock /Metal | Hip Hop |
| Walter Beasley | Blues / Folk | Jazz |
| Sevendust | Hip Hop | Rock / Metal |
| Big Bill roonzy | Reggae | Blues / Folk |
| Anita Baker | Rock | R&B / Soul |
| Cassandra Wilson | R&B / Soul | Jazz |
| Notixx | Reggae | Electronic |

**TABLE 2**: A COMPARISON BETWEEN SVD AND ITEM2VEC ON GENRE CLASSIFICATION TASK FOR VARIOUS SIZES OF TOP POPULAR ARTIST SETS

| Top (q) popular artists | SVD accuracy | item2vec accuracy |
|---|---|---|
| 2.5k | 85% | **86.4%** |
| 5k | 83.4% | **84.2%** |
| 10k | 80.2% | **82%** |
| 15k | 76.8% | **79.5%** |
| 20k | 73.8% | **77.9%** |
| 10k unpopular (see text) | 58.4% | **68%** |

provide a suggestion for the correct label using a simple k nearest neighbor (KNN) classifier.

In order to quantify the similarity quality, we tested the genre consistency between an item and its nearest neighbors. We do that by iterating over the top $q$ popular items (for various values of $q$) and check whether their genre is consistent with the genres of the $k$ nearest items that surround them. This is done by a simple majority voting. We ran the same experiment for different neighborhood sizes ($k = 6, 8, 10, 12$ and $16$) and no significant change in the results was observed.

**TABLE 3**: A QUALITATIVE COMPARISON BETWEEN ITEM2VEC AND SVD FOR SELECTED ITEMS FROM THE MUSIC DATASET

| Seed item (genre) | item2vec – Top 4 recommendations | SVD – Top 4 recommendations |
|---|---|---|
| David Guetta (Dance) | Avicii, Calvin Harris, Martin Solveig, Deorro | Brothers, The Blue Rose, JWJ, Akcent |
| Katy Perry (Pop) | Miley Cyrus, Kelly Clarkson, P!nk, Taylor Swift | Last Friday Night, Winx Club, Boots On Cats, Thaman S. |
| Dr. Dre (Hip Hop) | Game, Snoop Dogg, N.W.A, DMX | Jack The Smoker, Royal Goon, Hoova Slim, Man Power |
| Johnny Cash (Country) | Willie Nelson, Jerry Reed, Dolly Parton, Merle Haggard | Hank Williams, The Highwaymen, Johnny Horton, Hoyt Axton |
| Guns N' Roses (Rock) | Aerosmith, Ozzy Osbourne, Bon Jovi, AC/DC | Bon Jovi, Gilby Clarke, Def Leppard, Mtley Cre |
| Justin Timberlake (Pop) | Rihanna, Beyonce, The Black eyed Peas, Bruno Mars | JC Chasez. Jordan Knight, Shontelle, Nsync |

**TABLE 4**: A QUALITATIVE COMPARISON BETWEEN ITEM2VEC AND SVD FOR SELECTED ITEMS FROM THE STORE DATASET

| Seed item | item2vec – Top 4 recommendations | SVD – Top 4 recommendations |
|---|---|---|
| LEGO Emmet | LEGO Bad Cop, LEGO Simpsons: Bart, LEGO Ninjago, LEGO Scooby-Doo | Minecraft Foam, Disney Toy Box, Minecraft (Xbox One), Terraria (Xbox One) |
| Minecraft Lanyard | Minecraft Diamond Earrings, Minecraft Periodic Table, Minecraft Crafting Table, Minecraft Enderman Plush | Rabbids Invasion, Mortal Kombat, Minecraft Periodic Table |
| GoPro LCD Touch BacPac | GoPro Anti-Fog Inserts, GoPro The Frame Mount, GoPro Floaty Backdoor, GoPro 3-Way | Titanfall (Xbox One), GoPro The Frame Mount, Call of Duty (PC), Evolve (PC) |
| Surface Pro 4 Type Cover | UAG Surface Pro 4 Case, Zip Sleeve for Surface, Surface 65W Power Supply, Surface Pro 4 Screen Protection | Farming Simulator (PC), Dell 17 Gaming laptop, Bose Wireless Headphones, UAG Surface Pro 4 Case |
| Disney Baymax | Disney Maleficent, Disney Hiro, Disney Stich, Disney Marvel Super Heroes | Disney Stich, Mega Bloks Halo UNSC Firebase, LEGO Simpsons: Bart, Mega Bloks Halo UNSC Gungoose |
| Windows Server 2012 R2 | Windows Server Remote Desktop Services 1-User, Exchange Server 5-Client, Windows Server 5-User Client Access, Exchange Server 5-User Client Access | NBA Live (Xbox One) – 600 points Download Code, Windows 10 Home, Mega Bloks Halo Covenant Drone Outbreak, Mega Bloks Halo UNSC Vulture Gunship |

Table 2 presents the results obtained for $k = 8$. We observe that item2vec is consistently better than the SVD model, where the gap between the two keeps growing as $q$ increases. This might imply that item2vec produces a better representation for less popular items than the one produced by SVD, which is unsurprising since item2vec subsamples popular items and samples the negative examples according to their popularity.

We further validate this hypothesis by applying the same 'genre consistency' test to a subset of 10K unpopular items (the last row in Table 2). We define an unpopular item in case it has less than 15 users that played its corresponding artist. The accuracy obtained by item2vec was 68%, compared to 58.4% by SVD.

Qualitative comparisons between item2vec and SVD are presented in Tables 3-4 for Music and Store datasets, respectively. The tables present seed items and their 4 nearest neighbors (in the latent space). The main advantage of this comparison is that it enables the inspection of item similarities in higher resolutions than genres. Moreover, since the Store dataset lacks any informative tags / labels, a qualitative evaluation is inevitable. We observe that for both datasets, item2vec provides lists that are better related to the seed item than the ones that are provided by SVD. Furthermore, we see that even though the Store dataset contains weaker information, item2vec manages to infer item relations quite well.

## 5. CONCLUSION

In this paper, we proposed item2vec - a neural embedding algorithm for item-based collaborative filtering. item2vec is based on SGNS with minor modifications.

We present both quantitative and qualitative evaluations that demonstrate the effectiveness of item2vec when compared to a SVD-based item similarity model. We observed that item2vec produces a better representation for items than the one obtained by the baseline SVD model, where the gap between the two becomes more significant for unpopular items. We explain this by the fact that item2vec employs negative sampling together with subsampling of popular items.

In future we plan to investigate more complex CF models such as [1, 2, 3] and compare between them and item2vec. We will further explore Bayesian variants [12] of SG for the application of item similarity.